\documentclass{article}

\usepackage{microtype}
\usepackage{graphicx}
\usepackage{subcaption}
\usepackage{booktabs} % for professional tables

\usepackage{hyperref}

\usepackage[preprint]{icml2026}

\usepackage{amsmath}
\usepackage{amssymb}
\usepackage{mathtools}
\usepackage{amsthm}

\usepackage[capitalize,noabbrev]{cleveref}

\newcommand{\method}{VISTA\xspace}

\newcommand{\VLA}{\pi}
\newcommand{\backbone}{f}
\newcommand{\actExpert}{\phi}
\newcommand{\obs}{o}
\newcommand{\obsNoise}{\obs^\prime}
\newcommand{\obsTrack}{\tilde{\obs}}
\newcommand{\task}{g}
\newcommand{\taskTrack}{\tilde{\task}}
\newcommand{\act}{a}
\newcommand{\horizon}{H}
\newcommand{\timeStep}{t}
\newcommand{\outNum}{M}
\newcommand{\tokenPos}{\tau}
\newcommand{\traj}{\zeta}
\newcommand{\trajLen}{T}
\newcommand{\trajTrack}{\tilde{\zeta}}
\newcommand{\trackTotalNum}{N}
\newcommand{\trackNum}{n}
\newcommand{\batchID}{b}
\newcommand{\batchIDMax}{B}

\newcommand{\KL}{D_{\text{KL}}}
\newcommand{\diffTime}{k}
\newcommand{\diffTimeTotal}{K}
\newcommand{\diffVar}{\beta}

\newcommand{\Identity}{\mathbf{I}}

\DeclareMathOperator{\loss}{{\mathcal{L}}}
\DeclareMathOperator{\lossFunc}{\ell}
\DeclareMathOperator{\E}{\mathbb{E}}
\DeclareMathOperator{\D}{\mathcal{D}}
\DeclareMathOperator{\DTrack}{\tilde{\D}}

\newcommand{\distillWeight}{\gamma}
\newcommand{\DPOWeight}{\alpha}

\definecolor{tableHighlightColor}{HTML}{DEDEDE}

\definecolor{softgreen}{HTML}{08A045}
\definecolor{softred}{HTML}{D50032}
\newcommand{\tableGreenNote}[1]{\textcolor{softgreen}{{\footnotesize #1}}}
\newcommand{\tableRedNote}[1]{\textcolor{softred}{{\footnotesize #1}}}
\usepackage{amsmath,amsfonts,bm}
\usepackage[table]{xcolor}

\usepackage{xspace}
\usepackage{nicematrix}
\usepackage[flushleft]{threeparttable}
\usepackage{tablefootnote}
\usepackage{multirow}
\usepackage{hhline}
\usepackage{graphicx}
\usepackage{caption}
\usepackage{subcaption}
\usepackage{hanging}
\usepackage{color}
\usepackage{tikz}
\usetikzlibrary{calc,arrows,decorations.markings}
\usepackage{makecell, booktabs}
\usepackage{ifthen}
\usepackage{comment}
\usepackage{cancel}

\theoremstyle{plain}

\theoremstyle{definition}

\theoremstyle{remark}

\usepackage[textsize=tiny]{todonotes}

\icmltitlerunning{Submission and Formatting Instructions for ICML 2026}

\begin{document}

\twocolumn[
  \icmltitle{Enhancing Visual Conditioning via Track-Following Preference Optimization in Vision-Language-Action Models}

  % It is OKAY to include author information, even for blind submissions: the
  % style file will automatically remove it for you unless you've provided
  % the [accepted] option to the icml2026 package.

  % List of affiliations: The first argument should be a (short) identifier you
  % will use later to specify author affiliations Academic affiliations
  % should list Department, University, City, Region, Country Industry
  % affiliations should list Company, City, Region, Country

  % You can specify symbols, otherwise they are numbered in order. Ideally, you
  % should not use this facility. Affiliations will be numbered in order of
  % appearance and this is the preferred way.
  \icmlsetsymbol{equal}{*}
  \icmlsetsymbol{intern}{$\dagger$}
  \icmlsetsymbol{whileMS}{$\ddagger$}

  \begin{icmlauthorlist}
    \icmlauthor{Yiye Chen}{gt,intern}
    \icmlauthor{Yanan Jian}{nv,whileMS}
    \icmlauthor{Xiaoyi Dong}{ms}
    \icmlauthor{Shuxin Cao}{gt}
    \icmlauthor{Jing Wu}{os,intern}
    %\icmlauthor{}{sch}
    \icmlauthor{Patricio Vela}{gt}
    \icmlauthor{Benjamin E. Lundell}{arm,whileMS}
    \icmlauthor{Dongdong Chen}{ms}
    %\icmlauthor{}{sch}
    %\icmlauthor{}{sch}
  \end{icmlauthorlist}

  % \icmlaffiliation{intern}{Work done during an intern at Microsoft.}
  % \icmlaffiliation{whileMS}{Work done while working at Microsoft.}
  \icmlaffiliation{gt}{Georgia Tech}
  \icmlaffiliation{nv}{Nvidia}
  \icmlaffiliation{ms}{Microsoft}
  \icmlaffiliation{os}{University of Oxford}
  \icmlaffiliation{arm}{ARM}

  \icmlcorrespondingauthor{Yiye Chen}{yychen2019@gatech.edu}
  % \icmlcorrespondingauthor{Firstname2 Lastname2}{first2.last2@www.uk}

  % You may provide any keywords that you find helpful for describing your
  % paper; these are used to populate the "keywords" metadata in the PDF but
  % will not be shown in the document
  \icmlkeywords{Machine Learning, ICML}

  \vskip 0.3in
]

% this must go after the closing bracket ] following \twocolumn[ ...

% This command actually creates the footnote in the first column listing the
% affiliations and the copyright notice. The command takes one argument, which
% is text to display at the start of the footnote. The \icmlEqualContribution
% command is standard text for equal contribution. Remove it (just {}) if you
% do not need this facility.

% Use ONE of the following lines. DO NOT remove the command.
% If you have no special notice, KEEP empty braces:
% \printAffiliationsAndNotice{}  % no special notice (required even if empty)
% Or, if applicable, use the standard equal contribution text:
% \printAffiliationsAndNotice{\icmlEqualContribution}
\printAffiliationsAndNotice{
% \icmlEqualContribution
\textsuperscript{$\dagger$}Work done during an internship at Microsoft.
\textsuperscript{$\ddagger$}Work done while working at Microsoft.
}

\begin{abstract}
% VLA good, and naive VLA adds action token
Vision–Language–Action (VLA) models have demonstrated strong performance across a wide range of robotic manipulation tasks.
% Vision–Language–Action (VLA) models have demonstrated strong performance on a wide range of robotic manipulation tasks by extending pretrained Vision–Language Models (VLMs) with action generation capabilities.
% A common autoregressive VLA formulation augments the vocabulary of VLMs with action tokens, allowing the model to inherit rich world knowledge.
% But it leads to potential problem of misalignment and weak conditioning.
% Despite the success, introducing unseen action tokens can induce misalignment between perception and action modalities, causing the model to underutilize visual information during action prediction.
% Despite the success, the incorporation of additional training and structure modification for action predictions tasks might induce misalignment between perception and action modalities, leading to underutilization of visual information.
Despite the success, extending large pretrained Vision-Language Models (VLMs) to the action space can induce vision-action misalignment, where action predictions exhibit weak dependence on the current visual state, leading to unreliable action outputs.
% We show correlation between visual conditioning and performance
In this work, we study VLA models through the lens of visual conditioning and empirically show that successful rollouts consistently exhibit stronger visual dependence than failed ones.
% propose a method 
Motivated by this observation, we propose a training framework that explicitly strengthens visual conditioning in VLA models.
Our approach first aligns action prediction with visual input via preference optimization on a track-following surrogate task, 
% which encourages the model to leverage fine-grained visual cues to discriminate between subtly different action sequences.
and then transfers the enhanced alignment to instruction-following task through latent-space distillation during supervised finetuning.
% We then perform standard supervised instruction-following finetuning with latent distillation, which preserves the alignment in the model.
% Achieve better performance
Without introducing architectural modifications or additional data collection, our method improves both visual conditioning and task performance for discrete OpenVLA, and further yields consistent gains when extended to the continuous OpenVLA-OFT setting.
% Code: \url{https://anonymous.4open.science/r/ICML30025_submit-08F6}
Project website: \url{https://vista-vla.github.io/}.

% our approach consistently improves the performance of OpenVLA on both LIBERO and CALVIN benchmarks.
% % Extend to OFT receipe.
% Moreover, we show that the proposed framework naturally extends to the continuous action expert and parrallel-decoding setting,
% yielding significant gains for OpenVLA-OFT on the long-horizon CALVIN benchmark.
  
\end{abstract}

\section{Introduction}

% \red{My current version does not define "alignment" in the literature well, since this term is used in multiple way, which makes the motivation and idea ambiguous.
% For VLMs, the "preference alignment" is a training method, that can be used to address ungrounded generation / visual conditioning / hallucination issue".
% We study the problem under the same visual conditioning lens, and leverage track-action alignment to mitigate the issue.
% }

% \red{
% Also maybe should summarize how preference optimization address VLMs hallucination in Intro? or just mention in one sentence to give more context.
% I will refine the introduction and related works later.
% }

% \red{Use inverse dynamics rather than inverse kinematics}

% \red{Need to further categorize the recent VLAs with vision-language co-finetuning released in late 2025, my current summary is not clear enough.}

\begin{figure}[t!]
    \centering
    \vspace*{-0pt}
    \scalebox{0.9}{
    	\begin{tikzpicture}
         \node[anchor=north west] at (0in,0in)
          {{\includegraphics[width=1.0\linewidth,clip=true,trim=0pt 310pt 635pt 0pt]{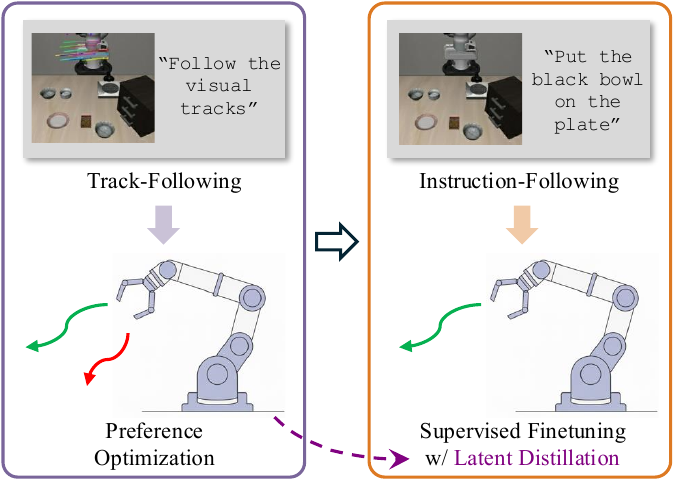}}};
        \end{tikzpicture}
    }
    % \vspace*{-15pt}
    \caption{
        \textbf{\method Overview.}
        We align VLA action outputs to visual tracks via preference optimization, 
        followed by supervised finetuning with latent distillation.
        Our experiments show that \method enhances visual conditioning and improves performance.
    }
    \label{fig:teaser}
    \vspace*{-13pt}
\end{figure}

% VLA, new action modules, misalignment, weak visual conditioning, weak results.
% Our method, enhance alignment, enhance visual conditioning, improve performance

% perhaps should cite VLM papers to justify why misalignment and visual conditioning are connected

% Figures 
%   The teaser figure should show the DPO-to-DistillSFT training paradigm.
%   The prelimiary study figure should show how ours improve both the visual conditioning and performance

% When summarizing results, can emphasize that our approach does not require any additional datasets, collecting more rollouts, or network structural change, etc.

% Introduce instruction following task, VLAs
Recent research has shown increasing interest in the important challenge of developing instruction-following robots capable of completing manipulation tasks conditioned on natural language commands\cite{CGNet, InnerMonologue, CLIPort, Perceiver-actor}.
Vision-Language-Action (VLA) models seek to address this challenge by extending large Vision-Language Models (VLMs) with action prediction modules. This paradigm seeks to transfer the rich perceptual and reasoning capabilities learned during VLM pretraining to embodied action prediction. 

Despite the current success, it remains unclear to what extent the world knowledge and visual understanding encoded in VLMs effectively propagates to the VLA.
Recent studies \cite{UP-VLA, embodiedR1} suggest that information extracted from VLMs may be insufficient for embodied decision-making, due to the discrepancy of the feature granularity required by the vision-language reasoning and visuomotor control.
Other work \cite{dontblindyourvla, KnowledgeInsulating} further shows the degradation in the visual representation due to finetuning on the extended action output domain.
These findings point to the potential \textit{misalignment} between perceptual understanding and action generation, wherein policies underutilize visual evidence and instead rely on spurious priors, resulting in unreliable action outputs.
To mitigate this issue, existing approaches introduce auxiliary tasks, such as 
video prediction \cite{GR1, VPP}, dynamics understanding \cite{univla, LAPA, moto}, spatial question answering \cite{pi0.5, chatvla}, and distillation from external experts \cite{VITA}, to enhance fine-grained visual understanding.
However, despite the incured substantial training cost, it remains unclear how those external knowledge translates to better action prediction, as their impact on the policy decision is hard to characterize.

% To preserve vision-language knowledge for action prediction, a large body of research is exploring co-training VLA on heterogeneous vision-language datasets, formulated as question answering, image understanding, or video prediction tasks. \red{cite a bunch}.
% Despite the scaled training cost, the VLA still might not effectively extrapolating from the vision-language co-training due to training stages and structural difference, where VLA and VLM tasks go through distinct paths \red{cite MOE type works} and experts \red{cite a bunck} within the model.
% Another thread of work that might mitigate the issue formulates the action tokenizer as a latent action model that comprehends the dynamics of the scene \red{cite}. While the latent action model guides the VLA training by annotating the action tokens, its dynamic knowledge might not transfer to VLA since the action model is independent of the VLA.

% What we did - visual conditioning
In this paper, we study the problem through the novel lens of \textit{visual conditioning}, which explicitly measures the extent of the impact of visual clue on the action prediction.
Intuitively, stronger visual conditioning is desirable for VLA models, as weak conditioning might indicate insufficient consideration of the current state and poor utilization of the VLM knowledge, leading to brittle action outputs.
To consolidate this notion, we quantify visual conditioning by measuring the shift in the predicted action distribution under the distortion of the visual input.
Under this formulation, we empirically demonstrate on the state-of-the-art VLA that successful rollouts exhibit stronger visual conditioning than failure ones, which validates the significance of visual conditioning for robust action generation.

% Our method to train it  - inspired from VLM preference optimization to reduce hallucination.
Based on this observation, we aim to improve VLA performance by explicitly strengthening visual conditioning during action prediction. 
Our approach is motivated by the success of \textit{preference optimization} in mitigating VLMs hallucinations, where training on pairs of preferred and dispreferred responses encourages stronger reliance on visual evidence \cite{C-DPO, POVID, CLIP-DPO}.
However, extending preference optimization to instruction-following VLA policies is non-trivial, as constructing meaningful preference pairs requires dispreferred action chunks that are both in-distribution and less likely to accomplish the task.
This necessitates costly on-policy rollouts and expert-defined trajectory ranking \cite{GRAPE}, which limits scalability and flexibility.

To address this challenge, we propose \textit{\textbf{\method}: Improving Visuomotor Coordination by Aligning \textbf{Vis}ual \textbf{T}racks and \textbf{A}ctions in VLA}.
The core idea of \method is to leverage visual tracks to make vision–action correspondence explicit, thereby enabling principled construction of preference pairs without model rollouts.
As illusrated in Figure \ref{fig:teaser}, \method augments standard supervised fine-tuning (SFT) with two additional training stages.
In the first stage, we apply Direct Preference Optimization \cite{DPO} to the \textit{track-following} task, where the VLA model predicts actions based on track-annotated images.
The track-following data are readily constructed from the instruction-following dataset by annotating the image with pixel tracks and applying minimal prompt modifications.
Preference pairs are then constructed entirely offline via in-batch pairing, which treats action chunks from other samples within the same batch as dispreferred responses.
We show that the track-following DPO implicitly optimizes inverse dynamics understanding in the VLA policy and greatly improves visual conditioning during action prediction.
In the second stage, \method performs instruction-following SFT with feature-space distillation from the frozen track-following model, which leverages the correspondence between the two tasks to transfer the enhanced visual grounding to the target task.

Without introducing architectural changes, auxiliary datasets, or on-policy rollouts, \method consistently improves visual conditioning and performance in the vanilla discrete autoregressive VLA model.
In particular, it improves OpenVLA \cite{openvla} by $3.15\%$ in average success rate on the LIBERO benchmark, demonstrating the benefit of strengthening visual conditioning through training.
Moreover, we also show that our method naturally extends to continuous parallel-decoding architectures by modifying only the final training stage.
This simple extension improves OpenVLA-OFT \cite{openvla-oft} on Calvin ABC$\rightarrow$D from $3.87$ to $4.02$ in average task completion count, corresponding to a $4\%$ relative improvement.

In summary, our contributions are:
\begin{itemize}
    \item We study Vision-Language-Action (VLA) models through the novel lens of \textit{visual conditioning} and demonstrate its importance for VLA performance.
    \item We introduce \method, a new training framework that improves visual conditioning in VLA policies by aligning action predictions to visual tracks.
    % without necessiating any network architectural change, additional training dataset, or expensive on-policy rollout collection.
    \item We show that \method enhances visual conditioning and consistently improves the performance of both discrete OpenVLA and continuous OpenVLA-OFT on standard benchmarks.
\end{itemize}

\section{Related Works}

\paragraph{Vision-Language-Action Models}
Early work in robot learning leverages vision and language foundation models as modular components within larger systems \cite{unisim, sayplan, sg2, voxposer}. 
More recent efforts instead focus on endowing large pretrained Vision–Language Models (VLMs) with end-to-end action prediction capabilities.
By tokenizing actions through discretization \cite{RT1, openvla} or frequency decomposition \cite{pi-fast}, 
the vanilla discrete autoregressive VLA models achieve strong performance powered by large-scale action-labeled datasets \cite{heterogeneous, pi0.5}.
Moderate architectural extensions, such as continuous-action experts \cite{openvla-oft, vla-adapter, robovlms} and parallel-decoding schemes \cite{openvla-oft, CoT-VLA}, have further improved the model expressivity and performance.

Despite the progress, recent studies report lack of fine-grained visual feature in VLM for embodied tasks \cite{UP-VLA, VLM2VLA, embodiedR1, spatialVLM} or degraded multimodal representation during VLA finetuning \cite{dontblindyourvla, VITA, KnowledgeInsulating}.
% raising concerns on the effectiveness of VLM-to-VLA paradigm.
A broad line of work seeks to address the issue by injecting world knowledge into VLAs,
through the task of latent action modeling \cite{moto, LAPA, univla}, embodied vision-language reasoning \cite{chatvla, ECoT}, dynamic modeling \cite{UnifiedVLA, UnifiedVideoActionModel, motus}, or video prediction \cite{CoT-VLA, CLOVER, SuSIE, GR1}. 
However, these approaches rely on auxiliary objectives whose impact on instruction-following action prediction is indirect and difficult to quantify, especially given separate training stages or distinct internal computation paths.
In contrast, we explicitly quantifies visual clue utilization in action generation, and propose a training method to enhance visual grounding.

\paragraph{Visual Track Guidance in Robot Learning}
Visual tracks, which depict pixel-level trajectories, have been widely used as a general representation of scene dynamics across diverse applications \cite{BootsTAPNext, motionPrompt}.
In robotics, visual track has facilitated policy training by serving as intermediate motion representations \cite{ATM, track2act} and enabling the extraction of dynamic knowledge from generic video data \cite{motionTracks}.
In the context of VLA models,
LLARVA \cite{llarva} employs future track prediction as an auxiliary objective, 
while TraceVLA \cite{traceVLA} uses visual tracks to represent historical action context. 
In contrast, we leverage visual tracks to formulate a preference optimization objective to align action prediction with visual input.

\paragraph{Multimodal Alignment and Preference Optimization}
Multimodal misalignment is a central challenge for large multimodal models. In Vision Language Models (VLMs), such misalignment often manifests as underutilization of visual input, leading to hallucinated output \cite{surveyVLMAlign, mitigateHallucination}.
Preference optimization has emerged as an effective strategy to address the issue, in which the models are trained on paired preferred and dispreferred responses for stronger visual grounding \cite{POVID, HA-DPO, CLIP-DPO, CSR}. 
In this paper, we study VLAs through the same lens, by showing the connection between visual conditioning and action prediction performance. Building on this insight, we introduce a preference optimization framework that aligns action prediction with visual input, strengthening visual grounding and improving downstream performance.

\section{Visual Conditioning in VLA Models}

\subsection{VLA Preliminaries}
\paragraph{Problem Formulation}
We consider the problem of building a VLA model $\VLA_\theta$, parameterized by $\theta$, which processes a language instruction $\task$ and a third-person RGB observation $\obs_\timeStep$ at time step $\timeStep$, and outputs a chunk of $\horizon$ actions $\act_{t:t+\horizon}$:
\begin{equation}
\label{eq:VLA}
    \act_{\timeStep: \timeStep+\horizon} = \VLA_\theta(\obs_\timeStep, \task)
\end{equation}
Although some work incorporates additional inputs such as robot proprioception and wrist-camera images, we exclude those modalities in all experiments for fair comparison.

\paragraph{Standard VLA Structures}
The VLA model $\VLA$ consists of a multimodal backbone $\backbone$ that encodes inputs into a set of $\outNum$ latents, followed by an action expert $\actExpert$ that maps these latents to the actions.
Two main categories of $\actExpert$ involve: 
Existing VLA approaches broadly fall into two representative categories based on the design of attention in $\backbone$ and architecture of $\actExpert$:
(1) Vainlla \textit{autoregressive discrete} formulation, which inherits the autoregressive backbone and token classification head from VLMs and outputs a distribution over a set of discrete action tokens: $\act_{\timeStep: \timeStep+\horizon} \sim \VLA(\cdot | \obs_\timeStep, \task)$;
% \red{may need to flesh out the autoregressive formulation here?}
(2) Extended \textit{parallel-decoding continuous} formulation, which adopts parallel-decoding attention and replaces the token head with a regression expert from the scratch that directly outputs the actions: $\act_{\timeStep: \timeStep+\horizon} = \VLA(\obs_\timeStep, \task)$. 
% Recent work \cite{pi0.5, univla} also show that multi-stage training involving both structures achieves the most superior results, where the VLA is pre-trained with a discrete expert to extract vision-language knowledge, followed by finetuning with a continuous expert for higher precision. 

In this work, we study training method under both formulations through OpenVLA \cite{openvla} and OpenVLA-OFT \cite{openvla-oft} architectures.
In both cases, $\backbone$ is finetuned from Prismatic VLM \cite{prismatic}, and each output latent corresponds to one action dimension, leading to $\outNum = 7 \horizon$ for a standard 7-DoF action space (translation, rotation, gripper).
OpenVLA instantiates the autoregressive discrete formulation,
whereas OpenVLA-OFT realizes the parallel-decoding continuous scheme, where $\actExpert$ is implemented as a 4-layer MLP network.

\paragraph{Instruction-following Dataset}
Training of a VLA requires an instruction-following dataset $\D = \{ (\traj, \task) \}$, where the trajectory $\traj$ records a sequence of actions and observations to accomplish the task $\task$ given initial observation $\obs_1$: $\traj = \{(\obs_t, \act_t)\}_{t=1}^\trajLen$. 
The trajectories are further segmented into action chunks of fixed length $\horizon$ via sliding window, which supervises the outputs of VLA formulated in Eq. \ref{eq:VLA}.

\subsection{Quantifying Visual Conditioning}
\label{sec:VCDefine}
% Inspired by Visual Contrastive Decoding (VCD) \cite{VCD}, which leverages visual perturbation to amplify visual impact in the response of VLMs, 
% we probe visual conditioning in the autoregressive discrete VLA by measuring the sensitivity of action predictions to visual input.
% \red{Shorten this one to: Inspired by Visual Contrastive Decoding (VCD) \cite{VCD},
% we probe visual conditioning in the autoregressive discrete VLA by measuring the sensitivity of action predictions to visual input.}
Inspired by Visual Contrastive Decoding (VCD) \cite{VCD},
we probe visual conditioning in the autoregressive discrete VLA by measuring the sensitivity of action predictions to visual input.
Formally, we define the visual conditioning at token position $\tokenPos$ as the KL-divergence between the predicted action token distributions conditioned on the original and perturbed observations, respectively:
\begin{equation}
\label{eq:VC}
\begin{aligned}
    \KL(\VLA(\cdot | \obs_\timeStep, \task, & \act_{\timeStep}^{<\tokenPos}) || \VLA(\cdot | \obsNoise_\timeStep, \task, \act_{\timeStep}^{<\tokenPos})) \\ 
    % \quad
    &\act_{\timeStep}^{<\tokenPos} \sim \VLA(\cdot | \obs_\timeStep, \task),
    \quad \tau = 1, \dots, \outNum
\end{aligned}
\end{equation}
where $\act_{\timeStep}^{<\tokenPos}$ are predicted tokens prior to position $\tokenPos$. $\obsNoise_\timeStep$ is the distorted image based on $\obs_\timeStep$. We follow VCD to implement the distortion as the forward diffusion process \cite{Diffusion}, by repetitively shrinking the original image and adding Gaussian noise for $\diffTimeTotal$ step. Equivalently, $\obsNoise_\timeStep$ is sampled from the following distribution: 
\begin{equation}
\begin{aligned}
    & q(\obs_\timeStep^{\diffTime} | \obs_\timeStep^{\diffTime-1}) 
    = 
    \mathcal{N}(\obs_\timeStep^{\diffTime} ; \sqrt{1-\diffVar_\diffTime} \obs_\timeStep^{\diffTime-1}, \diffVar_\diffTime \Identity), 
    \\
    & q(\obsNoise_\timeStep | \obs_\timeStep) = \prod_{\diffTime=1}^{\diffTimeTotal} q(\obs_\timeStep^{\diffTime} | \obs_\timeStep^{\diffTime-1}),
    \quad \obs_\timeStep^{0} = \obs_\timeStep
\end{aligned}
\end{equation}
where $\Identity$ refers to an identity matrix. We follow VCD to set $\diffTimeTotal = 999$ and schedule the variance linearly from $\diffVar_1=10^{-5}$ to $\diffVar_\diffTimeTotal=0.005$.

\subsection{Visual Conditioning and VLA Performance}

We begin by probing visual conditioning in OpenVLA finetuned on LIBERO-Spatial benchmark (Sec.~\ref{sec:ExpLIBERO}) with chunk size $\horizon=8$ (i.e. $\outNum=56$).
Using the definition in Eq.~\eqref{eq:VC}, we measure visual conditioning across all output tokens over 100 rollouts (10 tasks with 10 rollouts per task).
Rollouts are categorized into successful and failed groups based on simulation outcomes, and we report the mean and standard deviation aggregated over time steps and rollouts.

\begin{figure}[t!]
    \centering
    \vspace*{-0pt}
    \scalebox{0.95}{
    	\begin{tikzpicture}
         \node[anchor=north west] at (0in,0in)
          {{\includegraphics[width=1.0\linewidth,clip=true,trim=2pt 0pt 0pt 0pt]{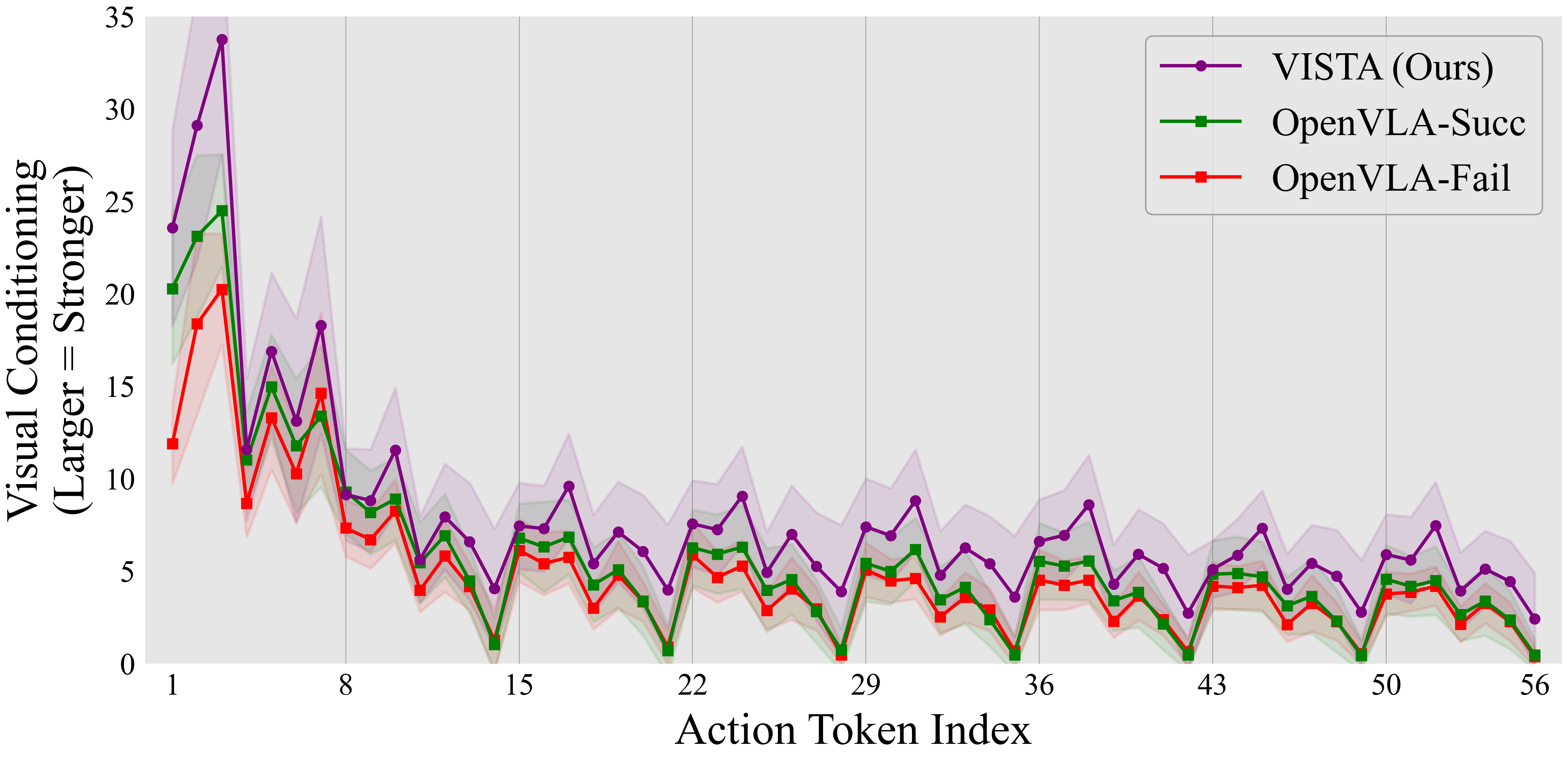}}};
        \end{tikzpicture}
    }
    \vspace*{-15pt}
    \caption{
        \textbf{Visual Conditioning of the 8-step OpenVLA and \method (Ours) in LIBERO-Spatial}.
        The periodic vertical grids indicate that each block of seven tokens decodes to a single action, leading to 56 output tokens for 8 actions.
        % \red{This figure needs to be on page 3!!!}
    }
    \label{fig:VCStudy}
    \vspace*{-15pt}
\end{figure}

As shown in Figure~\ref{fig:VCStudy}, successful rollouts consistently exhibit stronger visual conditioning than failed ones. 
This gap is most pronounced for early action tokens, which depend more heavily on visual input due to limited preceding token predictions.
These results establish a clear correlation between visual conditioning and VLA performance, which supports our notion that effective VLA policy requires strong visual grounding.
Motivated by the observation, we design a training approach to explicitly enhance vision-action alignment in Sec.~\ref{sec:method}.
As previewed in Figure~\ref{fig:VCStudy}, our method increases visual conditioning in OpenVLA, and leads to consistent performance improvement across multiple benchmarks as discussed in Sec.~\ref{sec:Exp}.

%%%%% figure %%%%%%%%%%%%%%%%%%%%%%%%%%%%%%%%%%%%%%%%%%%%%%%%%
\begin{figure*}[t!]
  % \vspace*{-0.1in}
  \centering
  \scalebox{0.90}{
    \begin{tikzpicture}
     \node[anchor=north west] at (0in,0in)
      {{\includegraphics[width=1.0\textwidth,clip=true,trim=0 210 260 0]{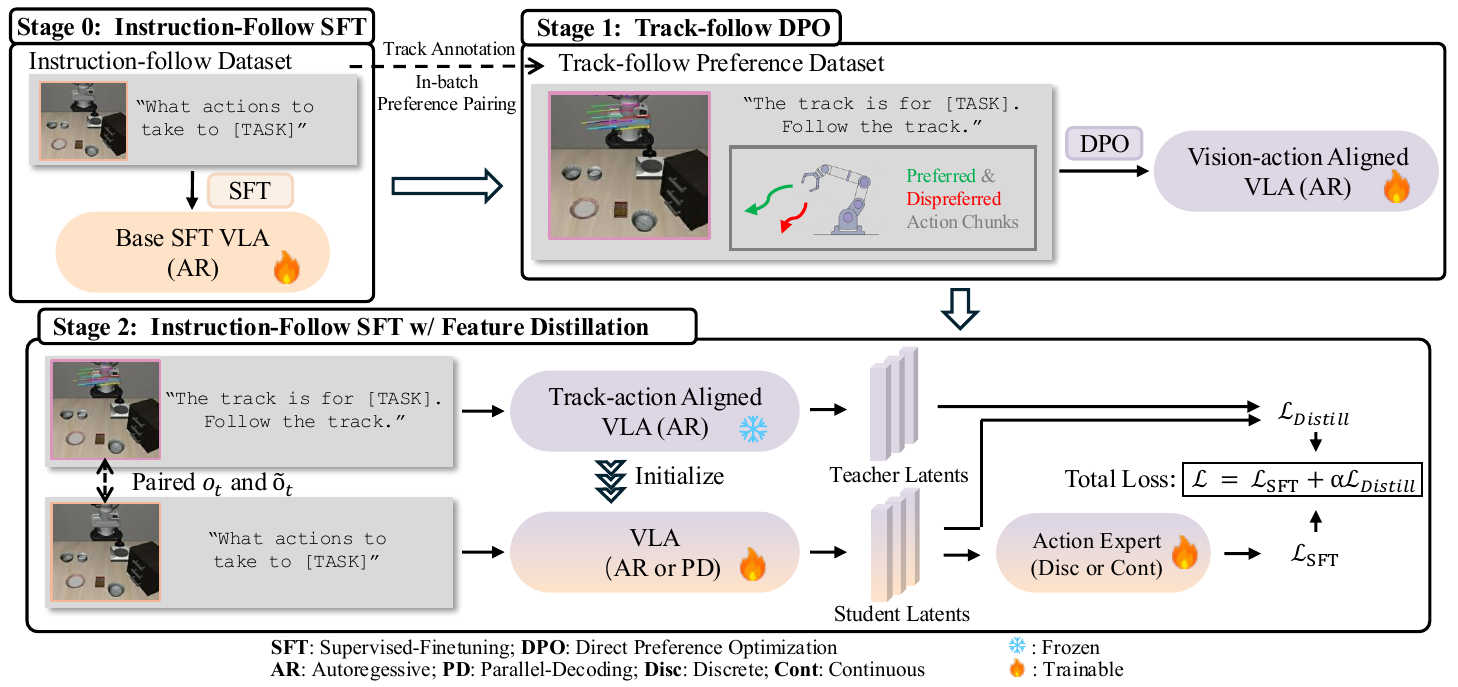}}};
    \end{tikzpicture}
  }
  % \vspace*{-0.05in}
   \vspace{-5pt}
  \caption{
    \textbf{\method Methodology}.
    Starting from a vanilla instruction-following SFT model,
    we apply DPO on track-following preference samples constructed from the instruction-following dataset to align action prediction with visual input.
    We then transfer the alignment to the instruction-following policy via latent distillation,
    resulting in enhanced visual conditioning and VLA performance.
  }
 \label{fig:Method}
 \vspace{-10pt}
\end{figure*}
%%%%%%%%%%%%%%%%%%%%%%%%%%%%%%%%%%%%%%%%%%%%%%%%%%%%%%%%%%%%%

\section{Methodology}
\label{sec:method}
As illusrated in Figure~\ref{fig:Method}, \method involves 3 training stages.
In Stage~0, we begin with standard instruction-following supervised fine-tuning (SFT) on a discrete autoregressive VLA, which provides a base model for the subsequent stages.
In Stage~1 (Sec.~\ref{sec:methodStage1}), we produce track-following data from the instruction-following dataset $\D$ and apply Direct Preference Optimization (DPO) to enhance vision-action alignment.
In Stage~2 (Sec.~\ref{sec:methodStage2}), we propagate the enhanced alignment to the instruction-following task of interest by additional SFT with latent distillation from the frozen copy of track-aligned VLA.

\subsection{Track-Follow Preference Optimization}
\label{sec:methodStage1}

\subsubsection{Track-following DPO Design}
\paragraph{Track-following Preference Dataset Construction.}
This stage starts from a base instruction-following VLA model to improve its visual conditioning via preference optimization.
To avoid the need for costly on-policy rollouts and expert-defined rewards when defining preference pairs for instruction-following tasks \cite{GRAPE}, 
we instead formulate preference optimization for the \textit{track-following} task, where the action prediction is guided by the annotated tracks on the image.
These tracks establish deterministic correspondence between visual input and action chunks output, enabling principled preference construction directly from the offline instruction-following dataset $\D$.

We first construct a track-following dataset $\tilde{\D}$ by annotating $\D$.
Specifically, for each time step $t$ within trajectory $\traj$, we employ an off-the-shelf point tracking model to track $\trackTotalNum$ points over the next $\horizon$ frames $(\obs_{t}, \obs_{t+1}, \dots, \obs_{t+min(t+\horizon-1, \trajLen}))$.
Following TraceVLA \cite{traceVLA} we retain only the active points by filtering based on L1 moving distances, and then randomly sample $\trackNum$ active tracks.
These tracks are overlaid on the initial frame $\obs_t$ to produce a track-annotated image $\obsTrack_t$, which depicts both the current state and the future dynamics given the labeled actions.
By repeating the process for all frames in $\D$, we obtain the track-following dataset $\tilde{\D} = \{ (\trajTrack, \taskTrack) \}$, where the track-annotated trajectory $\trajTrack = \{(\obsTrack_t, \act_t)\}_{t=1}^\trajLen$, and $\taskTrack$ is the updated prompt as shown in Figure \ref{fig:Method}.

We then construct track-following preference pairs on $\tilde{\D}$ via \textit{in-batch preference pairing}. 
Since visual tracks deterministically specify the preferred actions, action chunks drawn from other batch samples can naturally serve as in-distribution yet visuallly misaligned responses.
Concretely, given a randomly shuffled track-following batch $\{(\obsTrack_t^\batchID, \taskTrack^\batchID, \act^\batchID_{t:t+\horizon})\}_{\batchID=1}^\batchIDMax$, where $\batchIDMax$ is set as even,
we create the preference batch $\{(\obsTrack_t^\batchID, \taskTrack^\batchID, \act^\batchID_{w}, \act^\batchID_{l})\}_{\batchID=1}^\batchIDMax$,
where the preferred $\act^\batchID_{w}$ and dispreferred $\act^\batchID_{l}$ are set as:

\begin{equation}
    \begin{aligned}
    \act^\batchID_{w} = \act^\batchID_{t:t+\horizon}, \qquad
    \act^\batchID_{l} = 
        \begin{cases} 
            \act^{\batchID+1}_{t:t+\horizon}, & \text{if $\batchID$ is odd} \\
            \act^{\batchID-1}_{t:t+\horizon}, & \text{if $\batchID$ is even} 
        \end{cases}
    \end{aligned}
\end{equation}

\paragraph{Track-following DPO Formulation}
With the track-following preference data, we align action predictions to visual tracks via Direct Preference Optimization (DPO) \cite{DPO}.  
We optimize the objective:
\begin{equation}
\label{eq:DPO}
\begin{aligned}
    \loss&_{trackDPO}(\VLA_\theta) = - \E_{(\obsTrack, \taskTrack, \act_{w}, \act_{l})\sim \tilde{\D}} \\
        & \Big[ \log \sigma \Big( \DPOWeight \big(
        \log \frac{\VLA_\theta (\act_{w}; \obsTrack, \taskTrack)}{\VLA_{ref} (\act_{w}; \obsTrack, \taskTrack)} - 
        \log \frac{\VLA_\theta (\act_{l}; \obsTrack, \taskTrack)}{\VLA_{ref} (\act_{l}; \obsTrack, \taskTrack)}
        \big) \Big) \Big]
\end{aligned}
\end{equation}
where
$\sigma$ denotes the logistic function and $\DPOWeight$ controls the deviation from the reference policy $\VLA_{ref}$, which is fixed as the frozen Stage~0 SFT model.
Here, $\VLA (\act; \obsTrack, \taskTrack)$ is the conditional likelihood of an action chunk estimated by the autoregressive VLA model, which is calculated as the sequential product of the token-level probabilities from the model outputs.
By contrasting against preferred and dispreferred action chunks based on the visual track guidance, DPO encourages the policy to align predictions with visual evidence, thereby strengthening the visual conditioning.

\subsubsection{Track-following DPO implicitly trains inverse dynamics understanding}
Unlike prior approaches that explicitly learn inverse dynamics that recover actions based on the current and future states \cite{UniversalPolicy, Seer}, 
we show that our designed DPO program implicitly achieves this.  
As derived in \citet{DPO},
the DPO program in Eq. \ref{eq:DPO} implicitly optimizes the following RL objective:
\begin{equation}
\begin{aligned}
    \loss(\VLA_\theta) & =  - \E_{(\act, \obsTrack, \taskTrack) \sim \tilde{\D}} \\
    & r(\act, \obsTrack, \taskTrack)  + \KL[\VLA_\theta(\act ; \obsTrack, \taskTrack)) || \VLA_{ref}(\act ; \obsTrack, \taskTrack))]
\end{aligned}
\end{equation}

where the reward is estimated via the binary preference classification under the Bradley-Terry preference model \cite{BTModel}:
\begin{equation}
    \loss(r) = - \E_{(\obsTrack, \taskTrack, \act_{w}, \act_{l})\sim \tilde{\D}} 
        \big[ \log \sigma (r(\act_{w}, \obsTrack, \taskTrack) - r(\act_{l}, \obsTrack, \taskTrack))\big]
\end{equation}
% \begin{equation}
%     r(\act, \obsTrack, \taskTrack) = \DPOWeight \log \frac{\VLA_\theta(\act ; \obsTrack, \taskTrack))}{\VLA_{\text{ref}}(\act ; \obsTrack, \taskTrack)} + \DPOWeight \log Z (\obsTrack, \taskTrack)
% \end{equation}

% Since $\obsTrack$ encodes both the current visual state and future evolution through visual tracks, maximizing the reward encourages the model to increase the likelihood of the action chunks that are consistent with the visual motion, relative to the likelihood assigned by the reference instruction-following model $\pi_{\text{ref}}$.
% Together with the KL regularization that anchors the optimization to $\pi_{\text{ref}}$, this constrained RL program induces inverse dynamics understanding based on visual track input within the policy model.

Because preference pairs $(\act_{w}, \act_{l})$ are constructed from action chunks that are aligned or misaligned with $\obsTrack$, which encodes both the current visual state and future evolution through visual tracks,
the learned reward function evaluates how consistent an action chunk is with the observed visual motion.
Consequently, maximizing the reward, maximizing the reward encourages the policy to infer the action configurations encoded by visual tracks, thereby inducing inverse dynamics understanding within the policy model.

\subsection{Instruction-Following SFT with Latent Distillation}
\label{sec:methodStage2}

While the track-following DPO stage strengthens visual conditioning, the target instruction-following task does not provide visual track guidance at inference.
To transfer the improved vision–action alignment to instruction-following policies, we introduce a final training stage of SFT with latent-space distillation.

Specifically, given the aligned autoregressive VLA $\VLA_{\text{align}}$, we freeze it as a teacher model and initialize a trainable student $\VLA_\theta$ from its weights.
We then optimize $\VLA_\theta$ using the paired instruction and track-following $\D$ and $\tilde{\D}$, where each track-following sample is a track-annotated counterpart of an instruction-following sample.
The training objective:
\begin{equation}
\begin{aligned}
    \loss_{\text{SFT}} &= \E_{(\obs_t, \task, \act_{t:t+\horizon})\sim\D} \lossFunc_{SFT} (\act_{t:t+\horizon}, \VLA_\theta(\obs_t, \task)) \\
    \loss_{\text{Distill}} &= \E_{(\obs_t, \obsTrack_t, \task, \taskTrack, \act_{t:t+\horizon})\sim (\D, \DTrack)} \lossFunc_{\text{sim}} (\backbone_\theta(\obs_t, \task), \backbone_{\text{align}}(\obsTrack_t, \taskTrack))\\
    \loss_{\text{total}} &= \loss_{\text{SFT}} + \distillWeight \loss_{\text{Distill}}
\end{aligned}
\end{equation}
where $\lossFunc_{SFT}$ is standard SFT loss, and $\lossFunc_{sim}$ encourages the student to match the latent produced by the aligned teacher, thereby preserving the strengthened visual grounding.

Importantly, this distillation-based SFT framework allows the student $\VLA_\theta$ to differ from $\VLA_{\text{align}}$ in architecture. 
In our experiments, $\VLA_{\text{align}}$ is consistently instantiated as a discrete autoregressive OpenVLA model from prior stages, 
while $\VLA_\theta$ is either kept as the same or updated into a continuous parallel-decoding OpenVLA-OFT. 
Accordingly, $\lossFunc_{\text{SFT}}$ corresponds to the discrete next-token prediction loss or the continuous $L1$ regression loss. 
We empirically show that this procedure consistently improves performance in both settings, which indicates that the learned vision–action alignment transfers effectively across architectures.

\subsection{Implementation Details}
In this work, we use off-the-shelf BootsTAPNext \cite{BootsTAPNext} as the tracker.
We track $\trackTotalNum=900$ points on an uniform grid on the image and randomly sample $\trackNum=16$ active tracks for annotation.
The DPO parameter is fixed to $\DPOWeight=0.1$.
Unless otherwise specified, distillation loss $\lossFunc_{\text{sim}}$ is set as the negative mean cosine similarity. 
We denote the discrete and continuous instantiation of our method as \textbf{\method} and \textbf{\method-OFT}, respectively.
% corresponding to architectural choices in the final training stage.
For \method, all three training stages utilize OpenVLA-7B.
For \method-OFT, the first two stages use OpenVLA-7B, while the final stage switches to the  OpenVLA-OFT-7B.
% architecture with a L1 regression head initialized from the scratch.
We adopt LoRA \cite{LORA} with rank=32 in all training.
\section{Experiments}
\label{sec:Exp}

%%%%%%%%%%%%%%%%%%%%%%%%%%%%%%%%%%%%%%%%% CALVIN Results Table & Benchmark images.

\newcommand{\CalvinFirstColumnCell}[1]{\Block[l]{1-2}{\hspace{0.3em}#1}}
\begin{table*}[t!]
    \centering
    % \vspace{-4pt}
    % \small
    \begin{minipage}[t]{0.27\textwidth}
        \vspace{0pt}
        \centering
        \scalebox{0.88}{
            \includegraphics[width=1.0\textwidth,clip=true,trim=0 62 720 0]{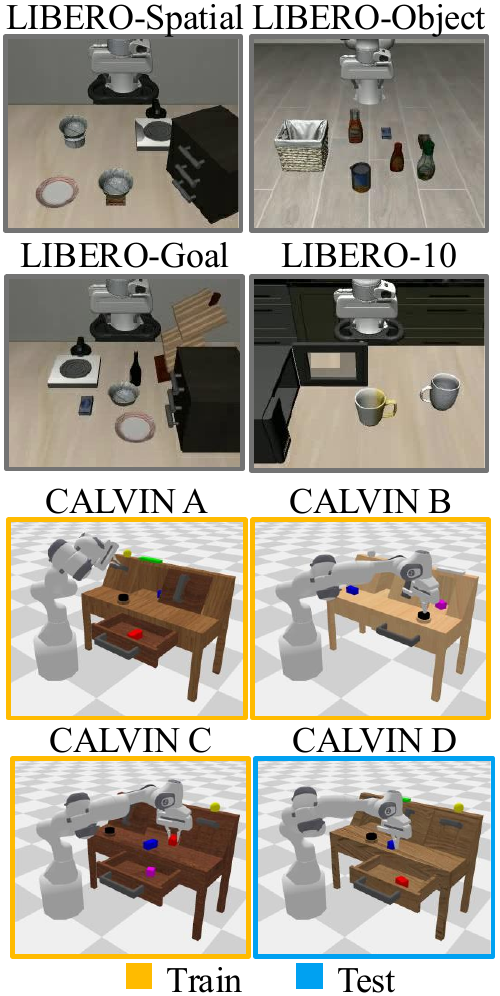}
        }
        \vspace{-4pt}
        \captionof{figure}{Illustration of benchmarks.}
        \label{fig:benchmarks}
    \end{minipage}\hfill
    \begin{minipage}[t]{0.71\textwidth}
        \vspace{0pt}
        \centering
        \setlength\extrarowheight{-0.5pt}
        \setlength{\tabcolsep}{10pt}
        \scalebox{1.0}{
            \begin{NiceTabular}{l @{\hspace{5.7em}} l c c c c c c}[
                vlines={3,8},
                % hlines
            ]
            % \hline
            \toprule[1.2pt]
                \Block{2-2}{Method} & &
                    \Block{1-5}{ Task completed in a row (\%) $\uparrow$} &&&&&
                    \Block{2-1}{\makecell[c]{Avg.\\Len.}} \\
                & & 1 & 2 & 3 & 4 & 5 & \\
            \midrule
            %
            %
            %   % Both implemented by RoboFlamingo. Both use third + wrist views.
                % Baselines are too old perhaps.
                % \multicolumn{2}{l|}{RT-1}  & 53.3 & 22.2 & 9.4 & 3.8 & 1.3 & 0.90 \\
                % \multicolumn{2}{l|}{RoboFlamingo}  & 82.4 & 61.9 & 46.6 & 33.1 & 23.5 & 2.48 \\
            %
            %
                \CalvinFirstColumnCell{SuSIE}  & & 87.0 & 69.0 & 49.0 & 38.0 & 26.0 & 2.69 \\
                \CalvinFirstColumnCell{GR-1} & & 85.4 & 71.2 & 59.6 & 49.7 & 40.1 & 3.06 \\
                
                \CalvinFirstColumnCell{VPP} & & 90.9 & 81.5 & 71.3 & 62.0 & 51.8 & 3.58\\
                \CalvinFirstColumnCell{DITA} & & 94.5 & 82.5 & 72.8 & 61.3 & 50.0 & 3.61\\
                
                \CalvinFirstColumnCell{CLOVER} & & 96.0 & 83.5 & 70.8 & 57.5 & 45.4 & 3.53 \\
                \CalvinFirstColumnCell{RoboDual} & & 94.4 & 82.7 & 72.1 & 62.4 & 54.4 & 3.66 \\
                \CalvinFirstColumnCell{RoboVLMs} & & 93.1 & 83.6 & 75.2 & 68.3 & 61.6 & 3.82\\
                \CalvinFirstColumnCell{UniVLA} & & 95.5 & 85.8 & 75.4 & 66.9 & 56.5 & 3.80\\
                \CalvinFirstColumnCell{ReconVLA} & & 95.6 & 87.6 & 76.9 & 69.3 & 64.1 & 3.95\\
        
            % \midrule
            \hline\hline
                \CalvinFirstColumnCell{OpenVLA$^{*}$} & & 92.8 & 78.3	 & 63.7	& 52.3	& 44.0 & 3.31 \\
                \rowcolor{tableHighlightColor}
                \CalvinFirstColumnCell{\method} & & 
                    \makecell[t]{93.5 \\ \tableGreenNote{(+0.7)}} & 
                    \makecell[t]{78.6 \\ \tableGreenNote{(+0.3)}}	& 
                    \makecell[t]{64.6 \\ \tableGreenNote{(+0.9)}}	& 
                    \makecell[t]{53.9 \\ \tableGreenNote{(+1.6)}}	& 
                    \makecell[t]{43.1 \\ \tableRedNote{(-0.9)}}    & 
                    \makecell[t]{3.34 \\ \tableGreenNote{(+0.03)}} \\
            \hline\hline
                \CalvinFirstColumnCell{OpenVLA-OFT$^{*}$} & & 92.1 & 83.2	 & 76.3	& 70.9	& 64.3 & 3.87 \\
                \rowcolor{tableHighlightColor}
                \CalvinFirstColumnCell{\method-OFT} & & 
                \makecell[t]{94.8 \\ \tableGreenNote{(+2.7)}} & 
                \makecell[t]{87.5 \\ \tableGreenNote{(+4.3)}} & 
                \makecell[t]{80.1 \\ \tableGreenNote{(+3.8)}} & 
                \makecell[t]{73.3 \\ \tableGreenNote{(+2.4)}} & 
                \makecell[t]{66.6 \\ \tableGreenNote{(+2.3)}} & 
                \makecell[t]{\textbf{4.02} \\ \tableGreenNote{(+0.15)}} \\
            % \hline
            \bottomrule[1.2pt]
            \end{NiceTabular}
        }
        \caption{\textbf{CALVIN ABC$\rightarrow$D Evaluation Results}. 
            \textcolor{softgreen}{Green} (\textcolor{softred}{red}) numbers indicate performance gains (drops) relative to the OpenVLA or OpenVLA-OFT counterpart.
            $^*$Results are reproduced
        }
        \label{tab:resultsCalvin}
    \end{minipage}
    \vspace{-12pt}
\end{table*}
%%%%%%%%%%%%%%%%%%%%%%%%%%%%%%%%%%%%%%%%%%%%%%%%%%%%%%%%%%%%%%%%%%%%%%%%%%%

%%%%%%%%%%%%%%%%%%%%%%%%%%%%%%%%%%%%%%%%%%%%%%%%%%%%%%% LIBERO Results Table
\begin{table}[t!]
    \centering
    % \small
    \setlength\tabcolsep{2.5pt}
    \setlength\extrarowheight{-0.5pt}
    \begin{NiceTabular}{l|cccc|c}[
            vlines = {2,6},     % vertical lines only before 2nd & 6th cell
            % hlines            % no horizontal lines
        ]
        \toprule[1.2pt]
        Methods & Spatial & Object & Goal & Long & Mean \\
        \hline
        Diffusion Policy & 78.3 & 92.5 & 68.3 & 50.5 & 72.4 \\
        Octo             & 78.9 & 85.7 & 84.6 & 51.1 & 75.1 \\
        LAPA             & 73.8 & 74.6 & 58.8 & 55.4 & 65.7 \\
        TraceVLA         & 84.6 & 85.2 & 75.1 & 54.1 & 74.8 \\
        SpatialVLA       & 88.2 & 89.9 & 78.6 & 55.5 & 78.1 \\
        % \hhline{=|====|=}
        \hline\hline
        OpenVLA-1Step    & 84.7 & 88.4 & 79.2 & 53.7 & 76.5 \\
        OpenVLA-8Step$^{*}$ & 79.6 & 88.6 & 82.4 & 54.4 & 76.3 \\
        
        \rowcolor{tableHighlightColor}
        VISTA (L2) & \makecell[t]{87.8\\ \tableGreenNote{(+8.2)}} 
                   & \makecell[t]{89.4\\ \tableGreenNote{(+0.8)}} 
                   & \makecell[t]{80.5\\ \tableRedNote{(-1.9)}} 
                   & \makecell[t]{58.6\\ \tableGreenNote{(+4.9)}} 
                   & \makecell[t]{\underline{79.1}\\ \tableGreenNote{(+2.6)}} \\
        
        \rowcolor{tableHighlightColor}
        VISTA & \makecell[t]{84.8\\ \tableGreenNote{(+5.2)}}
              & \makecell[t]{91.6\\ \tableGreenNote{(+3.0)}}
              & \makecell[t]{84.2\\ \tableGreenNote{(+1.8)}}
              & \makecell[t]{57.0\\ \tableGreenNote{(+2.6)}}
              & \makecell[t]{\textbf{79.4}\\ \tableGreenNote{(+3.1)}} \\
        \bottomrule[1.2pt]
    \end{NiceTabular}
    \caption{
        \textbf{LIBERO Evaluation Results.}
        % Success rates are evaluated over 500 episodes per suite.
        \method improves upon OpenVLA 
        % without any model architectural change or extra data, 
        and outperforms a range of baseline methods.
        % The default cosine similarity distillation loss achieves slightly superior performance compared to L2 loss.
        \textcolor{softgreen}{Green} (\textcolor{softred}{red}) numbers indicate performance gains (drops) relative to the OpenVLA-8Step counterpart.
        $^*$Results are reproduced.
    }
    \label{tab:ResultsLIBERO}
    \vspace{-20pt}
\end{table}
%%%%%%%%%%%%%%%%%%%%%%%%%%%%%%%%%%%%%%%%%%%%%%%%%%%%%%%%%%%%%%%%%%%%%%%%%%%

\subsection{LIBERO Benchmark Evaluation}
\subsubsection{Experimental Setups}
\label{sec:ExpLIBERO}
\paragraph{Evaluation Settings}
We first evaluate \method on the LIBERO manipulation benchmark \cite{LIBERO} to verify if our method improves visual conditioning and model performance. 
\method-OFT is excluded from this experiment and is evaluated only on the larger CALVIN benchmark (Sec.~\ref{sec:ExpCalvin}) due to its stronger expressivity.
As shown in Figure~\ref{fig:benchmarks}, LIBERO consists of four suites: Spatial, Object, Goal, and Long. Each suite contains 10 tasks, with 50 demonstrations per task for training.
We preprocess the data following OpenVLA \cite{openvla} to upscale the image resolution to $256\times256$, exclude failure trajectories, and remove static actions. 
Each method is evaluated under the success rate (SR) on 500 episodes per suite (10 tasks $\times$ 50 episodes per task).
The training configuration of \method is detailed in Appendix~\ref{sec:AppendixTrainDetail}.

% \paragraph{\method Training Configuration}

% \red{This info is already included in the appendix. Maybe remove this paragraph.}

% % We set the planning horizon to $\horizon=8$, leading to $\outNum=56$ tokens output per prediction in all three stages.\red{Put this to impelmentation details}
% The first SFT stage trains for 80K steps for LIBERO-Long and 60K steps, with a global batch size of 128.
% The second DPO stage trains for 60K steps for all suites with a global batch size of 16.
% The final SFT w/ feature distill stage trains for 20K steps for LIBERO-Spatial and LIBERO-Goal and 30K steps for LIBERO-Object and LIBERO-Long, with a global batch size of $96$.
% We further ablate the choice of distillation loss $\lossFunc_{\text{sim}}$ on LIBERO by replacing the negative cosine similarity with an L2 distance. The resulting variant is denoted as \textit{\method (L2)}.

\paragraph{Ablation on Distillation Function}
We further ablate the choice of distillation loss $\lossFunc_{\text{sim}}$ on LIBERO by replacing the negative cosine similarity with an L2 distance. The resulting variant is denoted as \textit{\method (L2)}.

\paragraph{Baselines}
We compare against OpenVLA \cite{openvla} trained with vanilla SFT under both $\horizon=8$ and its original $\horizon=1$. 
We additionally compare with TraceVLA \cite{traceVLA}, which adopts the same OpenVLA architecture but augments it with historical visual trace inputs.
Further state-of-the-art baselines include Diffusion Policy \cite{DiffusionPolicy}, LAPA \cite{LAPA}, Octo \cite{OCTO}, and SpatialVLA \cite{SpatialVLA}.
All methods are evaluated using only static third-person RGB observations.
%SpatialVLA  confirm this in the github issue: https://github.com/SpatialVLA/SpatialVLA/issues/21#issuecomment-2733202443

\subsubsection{Experimental Results}
For OpenVLA-8Step, \method,  and \method (L2), we report the results with Receding Horizon Control (RHC), executing only the first $4$ actions out of the predicted $\horizon=8$ before replanning. We found that RHC improves their performance in the relatively small LIBERO benchmark. 
Performance without RHC is reported in Appendix~\ref{sec:AppendixLIBEROResultsFull} 

\paragraph{Overall Results.}
The results are summarized in Table \ref{tab:ResultsLIBERO}. 
Without introducing any model architectural modifications or additional training data, \method consistently improves upon OpenVLA across all LIBERO suites, achieving an average SR gain of $3.1\%$.
Coupled with the improved visual conditioning shown in Figure \ref{fig:VCStudy}, 
these performance gains validate our training design and support our hypothesis that stronger visual conditioning benefits VLA performance.
Moreover, \method surpasses other baselines including TraceVLA, justifying the performance the gains from improved training recipe relative to the mere inclusion of track guidance. 
Overall, \method demonstrates competitive performance relative to all evaluated baselines.

\paragraph{Distillation Function Ablation Results.} 
% We examine the variants of distillation loss function $\lossFunc_{\text{sim}}$ in LIBERO, 
% testing L2 other than the default cosine similarity. 
As shown in Table \ref{tab:ResultsLIBERO}, 
\method (L2) variant also outperforms the baseline OpenVLA, which further demonstrate the effectiveness of our method regardless of the distillation loss instantiation.
However, compared to cosine similarity, the L2 variant achieves an average SR that is $0.5\%$ lower, with less consistent gains across task suites due to the $1.9\%$ decrease relative to OpenVLA on LIBERO-Goal.
Hence, we default to the cosine similarity in remaining experiments.

\subsection{Calvin Benchmark Evaluation}
\label{sec:ExpCalvin}
\subsubsection{Experimental Setups}
\paragraph{Evaluation Settings}
Next, we evaluate the generalization and long-horizon instruction-following capacities of VLA models on the CALVIN \cite{CALVIN} manipulation benchmark, which comprises 4 environments (A–D) as illustrated in Figure~\ref{fig:benchmarks}.
We focus on the most challenging ABC$\rightarrow$D setting,  where policies are trained on environments A–C and evaluated on D to test the generalization capacity.
Our training is conducted using only the subset annotated with language instruction, which includes $17,870$ episodes spanning 34 diverse tasks.
Evaluation follows the Long-Horizon Multi-Task Language Control protocol, under which the models are tested in 1000 sequences of five sequential tasks.
Within each sequence, the policy must solve each new task starting from the final state of the previous, which increases difficulty due to broader state-space coverage.
Performance is measured by per-task completion rate and the average task completion length.
We evaluate both \method and \method-OFT, with detailed training configurations listed in Appendix.~\ref{sec:AppendixTrainDetail}.

% \paragraph{\method Training Configuration}
% We test both \method and \method-OFT variants. 
% For both variants, stage 1 trains an 8-step OpenVLA for $200K$ steps with the global batch size of $128$, following the reproduction in UniVLA \cite{univla},
% and stage 2 performs track-following DPO on the same discrete architecture for $5$ epochs with the global batch size of $48$.
% For stage 3, we train \method for $12$ epochs with global batch size of $96$, and \method-OFT for $6$ epochs with the global batch size of $64 (8\times8)$ 

%%%%% Analysis & Ablation Figure %%%%%%%%%%%%%%%%%%%%%%%%%%%%
\begin{figure*}[t!]
  % \vspace*{-0.1in}
  \centering
  \scalebox{0.92}{
    \begin{tikzpicture}
     \node[anchor=north west] at (0in,0in)
      {{\includegraphics[width=1.0\textwidth,clip=true,trim=0
      290 195 0]{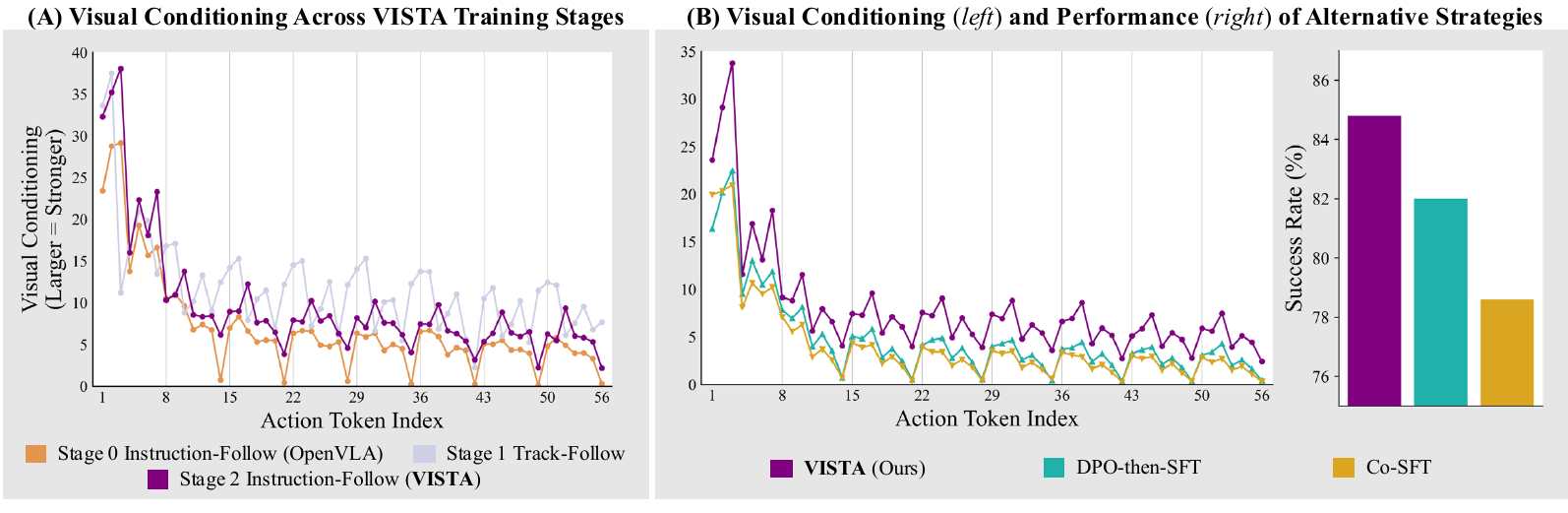}}};
    \end{tikzpicture}
  }
  % \vspace*{-0.05in}
  \caption{
    \textbf{Analysis and Ablation on LIBERO-Spatial}.
    \textbf{(A)} Change of visual conditioning during the \method training stages;
    \textbf{(B)} Ablation on the visual conditioning and VLA performance of alternative training strategies.
  }
 \label{fig:AnalysisAblate}
 \vspace{-12pt}
\end{figure*}
%%%%%%%%%%%%%%%%%%%%%%%%%%%%%%%%%%%%%%%%%%%%%%%%%%%%%%%%%%%%%

\paragraph{Baselines}

We compare \method and \method-OFT against OpenVLA and OpenVLA-OFT to highlight the advantage of the proposed training method.
We further benchmark diverse VLA designs, including RoboDual \cite{RoboDual}, which is a cooperation system of the OpenVLA and a trained specialist, and DITA \cite{DITA}, which is a generalist policy based on Diffusion Transformer (DiT).
Other than that, we benchmark a range of VLAs that incorporate additional action-free vision-language auxiliary tasks, including SuSIE \cite{SuSIE}, CLOVER \cite{CLOVER}, GR-1 \cite{GR1}, VPP \cite{VPP}, UniVLA \cite{univla}, and ReconVLA \cite{ReconVLA}.
Note that for RoboVLMs and VPP, we tabulate their best performance under the single static third-person view setting.

% RoboDual adds a specialist policy to cooperate with OpenVLA, which acts as a generalist. Use 3rd-view + gripper (specialist).

% CLOVER is a close-loop video prediction for error feedback + robotic policy.

% SuSIE predicts goal image and then actions, with 3rd view only.

% GR-1 first performs non-robotic video generative pre-training, then train to predict robotic action \& video. Use 3rd-view + gripper.

% ReconVLA add auxiliary gaze region reconstruction objective. 

% VPP builds on video prediction model, by first finetuning it to text-guided video prediction model on manipulation dataset, and then infer action with an explicit inverse dynamic model. Single-view results are in the appendix.  

% DITA is a Diffusion Transformer (DiT) generalist policy, using 3rd view only.

% RoboVLMs use the best variant under the 3rd-view-only setting, which uses Paligemma \cite{paligemma} VLM backbone.

\subsubsection{Experimental Results}
As shown in Table~\ref{tab:resultsCalvin}, our training method consistently improves both OpenVLA and OpenVLA-OFT in average length.
In particular, \method-OFT improves the task completion rate across all tasks of OpenVLA-OFT by more than $2.3\%$ and yields a $4\%$ relative increase in average completion length. 
This suggests that our designed track-following DPO extracts latents richer with vision and dynamic information, which guides better action generation.
What's more, despite relying on no external non-robotic dataset for additional pretraining or co-finetuning, \method-OFT surpasses a series of baselines in this category, and achieves the state-of-the-art performance in the CALVIN ABC$\rightarrow$D benchmark.
This further supports the efficacy of our training method designed to enhance visual conditioning.

\subsection{Analysis and Ablation}

\paragraph{Change of model visual conditioning during \method training.}
To examine how \method modulates visual conditioning in VLA models, we track its evolution across all training stages using the metric defined in Eq.~\ref{eq:VC}. 
Specifically, we measure the visual conditioning of the discrete \method at each stage on a 200 randomly sampled time steps from the LIBERO-Spatial training split.
Models from Stages~0 and~2 are evaluated on the instruction-following task with track-free images, while the Stage~1 model is evaluated on the track-following task with track-annotated observations.

Results are shown in Figure~\ref{fig:AnalysisAblate}(A).
The model trained solely with instruction-following SFT (Stage~0, equivalent to OpenVLA) exhibits the weakest visual conditioning, with near-zero values for some tokens, indicating complete ignorance of visual clue. 
Applying track-following DPO in Stage~1 substantially increases visual conditioning. 
Although the subsequent instruction-following training slightly reduces this effect, the final model retains markedly stronger visual conditioning than the Stage~0 baseline thanks to our distillation design.
These results support our claim that \method improves visual grounding of the VLA model,
which explains its performance gain compared to OpenVLA.

\paragraph{Ablation on alternative strategies.}
We further validate our design by comparing \method against two naive alternatives:
(1) \textit{DPO-then-SFT} applies track-following DPO followed by instruction-following SFT without latent distillation, analogous to a single iteration of DPO and SFT as in GRAPE \cite{GRAPE}. and serves to validate the efficacy of adding distillation loss $ \loss_{\text{Distill}}$ as in \method;
This baseline isolates the contribution of the distillation loss $\loss_{\text{Distill}}$. 
(2) \textit{Co-SFT},  performs a single-stage SFT on a mixture of instruction-following and track-following data, which resembles approaches that jointly train on action prediction and auxiliary world-modeling objectives.

Figure \ref{fig:AnalysisAblate}(B) shows visual conditioning and performance for all methods.
We observe the same ordering in both metrics: \method achieves the highest visual conditioning and performance, followed by DPO-then-SFT, and then Co-SFT.
This monotonic trend further justifies the clear correlation between visual conditioning and downstream VLA performance empirically.
Co-SFT yields the weakest visual conditioning and lowest performance, indicating that naive co-training with auxiliary objectives is insufficient for improving visual grounding. While DPO-then-SFT improves upon Co-SFT, it remains inferior to \method, underscoring the importance of latent distillation for effectively transferring visual grounding to instruction-following policies.

\section{Conclusion}

Our work examines the visual conditioning in VLA, which is quantified as the impact of visual clue on the output action distribution, and show its correlation with the VLA model performance.
Motivated by this finding, we further propose a VLA training recipe, named \method, that aligns the action prediction with the visual track guidance.
Our experiments show that \method facilitates the visual grounding, and consistently improves the VLA performance under both the naive discrete autoregressive and the continuous parallel-decoding (i.e. OFT) configurations. 

Due to compute constraints, our study focuses on the finetuning stage. 
However, we hope that this work can inspire future efforts on aligning action predictions with visual inputs in VLAs through extensive pretraining with mixed action-free video data, 
which can potentially enhances the performance and generalizability of VLA models.

% % ---------------- Acknowledgements should only appear in the accepted version.
% \section*{Acknowledgements}

% \textbf{Do not} include acknowledgements in the initial version of the paper
% submitted for blind review.

% If a paper is accepted, the final camera-ready version can (and usually should)
% include acknowledgements.  Such acknowledgements should be placed at the end of
% the section, in an unnumbered section that does not count towards the paper
% page limit. Typically, this will include thanks to reviewers who gave useful
% comments, to colleagues who contributed to the ideas, and to funding agencies
% and corporate sponsors that provided financial support.

% % ------------------------------------------------------------------

\section*{Impact Statement} % Does not count toward the paper page limit

%%%%%%%%%%%%%%%%%%%%%%%%%%% Guidance %%%%%%%%%%%%%%%%%%%%%%%%%%%%%%%%%%%%%
% Authors are \textbf{required} to include a statement of the potential broader
% impact of their work, including its ethical aspects and future societal
% consequences. This statement should be in an unnumbered section at the end of
% the paper (co-located with Acknowledgements -- the two may appear in either
% order, but both must be before References), and does not count toward the paper
% page limit. In many cases, where the ethical impacts and expected societal
% implications are those that are well established when advancing the field of
% Machine Learning, substantial discussion is not required, and a simple
% statement such as the following will suffice:

% ``This paper presents work whose goal is to advance the field of Machine
% Learning. There are many potential societal consequences of our work, none
% which we feel must be specifically highlighted here.''

% The above statement can be used verbatim in such cases, but we encourage
% authors to think about whether there is content which does warrant further
% discussion, as this statement will be apparent if the paper is later flagged
% for ethics review.
%%%%%%%%%%%%%%%%%%%%%%%%%%%%%%%%%%%%%%%%%%%%%%%%%%%%%%%%%%%%%%%%%%%%%%%%%

Our work aims to advance command-following robotic manipulation, enabling robots to physically interact with the environment in response to human instructions. 
Such capabilities have the potential to improve efficiency across many societal activities. 
By enhancing the visual dynamic understanding in a robot, our method represents a step toward more robust and reliable manipulation systems.

% In the unusual situation where you want a paper to appear in the
% references without citing it in the main text, use \nocite
% \nocite{langley00}

\bibliography{reference}
\bibliographystyle{icml2026}

%%%%%%%%%%%%%%%%%%%%%%%%%%%%%%%%%%%%%%%%%%%%%%%%%%%%%%%%%%%%%%%%%%%%%%%%%%%%%%%
%%%%%%%%%%%%%%%%%%%%%%%%%%%%%%%%%%%%%%%%%%%%%%%%%%%%%%%%%%%%%%%%%%%%%%%%%%%%%%%
% APPENDIX
%%%%%%%%%%%%%%%%%%%%%%%%%%%%%%%%%%%%%%%%%%%%%%%%%%%%%%%%%%%%%%%%%%%%%%%%%%%%%%%
%%%%%%%%%%%%%%%%%%%%%%%%%%%%%%%%%%%%%%%%%%%%%%%%%%%%%%%%%%%%%%%%%%%%%%%%%%%%%%%
\newpage
\appendix
\onecolumn  % can be removed

% \section{Code submission}
% Please see the anonymous link: 

% \url{https://anonymous.4open.science/r/ICML30025_submit-08F6}

\section{Additional Implementation Details}

\subsection{Visual Track Generation Details}
\label{sec:AppendixTrackDetail}
This section provides additional details on visual track generation.
We adopt different strategies for LIBERO and CALVIN benchmarks.
For LIBERO, we follow the protocol in TraceVLA \cite{traceVLA} to track over overlapping video segments. Specifically, each demonstration is divided into overlapping segments of 32 size (e.g. [1, 32], [17, 48], etc). 
For each segment, we initialize $\trackTotalNum = 900$ tracking points in the first frame of the segment and track their coordinates throughout the entire 32-frame window. With this design, any track whose starting frame falls within the first 16 frames of a segment is guaranteed to have a future trajectory of at least 16 frames.
This design also enables the inclusion of newly visible regions and increases diversity of visual tracks across segments.
For Calvin, since each language-annotated demonstration has a maximum length of 64 frames, we directly track the initialized points over the entire trajectory without further segmentation.

To btain valid and active visual tracks, we apply several filtering criteria.
First, we remove static points by calculating frame-to-frame pixel displacement over the tracking interval (32 frames for LIBERO and up to 64 frames for CALVIN). Points with an average displacement below 2 pixels are discarded. 
Next, we remove points that exhibit insufficient temporal motion consistency, defined as having displacements greater than 2 pixels in fewer than one quarter of the tracked frames. Finally, we discard points that exhibit abrupt motion, specifically those with a displacement exceeding 10 pixels between any pair of consecutive frames. This filtering strategy effectively removes false-positive tracks, particularly those with spurious jumps.

\begin{figure*}[h!]
  % \vspace*{-0.1in}
  \centering
  \scalebox{0.95}{
    \begin{tikzpicture}
     \node[anchor=north west] at (0in,0in)
      {{\includegraphics[width=1.0\textwidth,clip=true,trim=0
      235 0 0]{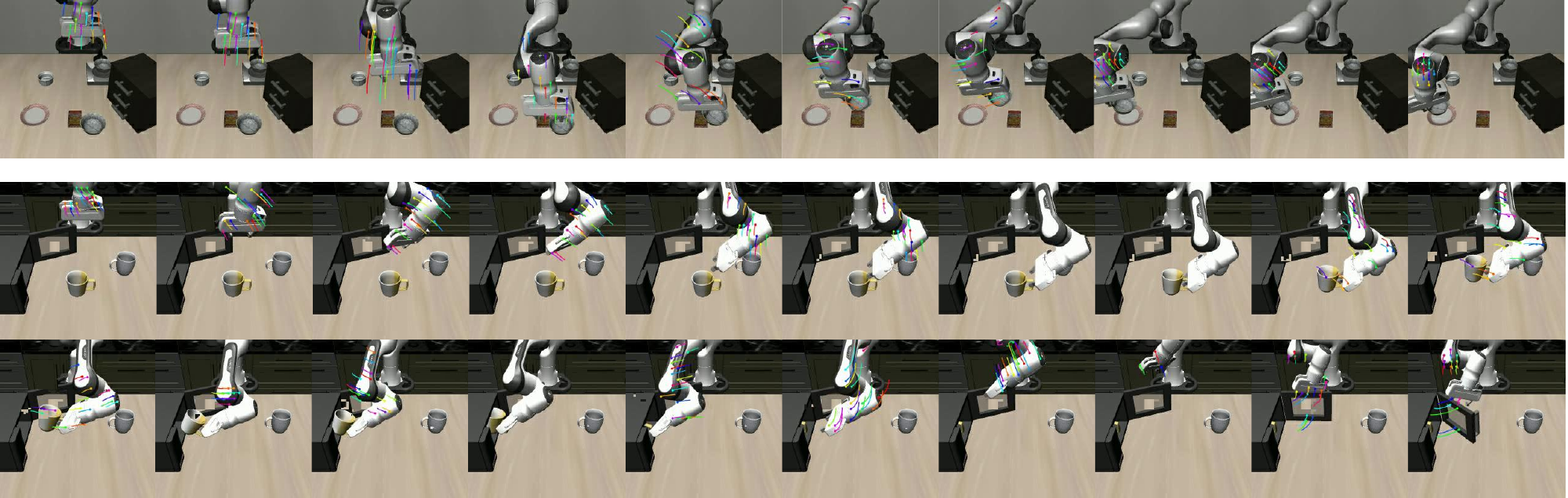}}};
    \end{tikzpicture}
  }
  % \vspace*{-0.05in}
  \caption{
    \textbf{Track annotation results on LIBERO dataset}. Top: LIBERO-Spatial; Bottom: LIBERO-Long.
    \textit{Best viewed when zoomed in!}
  }
 \label{fig:appTrackLibero}
\end{figure*}

\begin{figure*}[h!]
  % \vspace*{-0.1in}
  \centering
  \scalebox{0.95}{
    \begin{tikzpicture}
     \node[anchor=north west] at (0in,0in)
      {{\includegraphics[width=1.0\textwidth,clip=true,trim=0
      340 0 0]{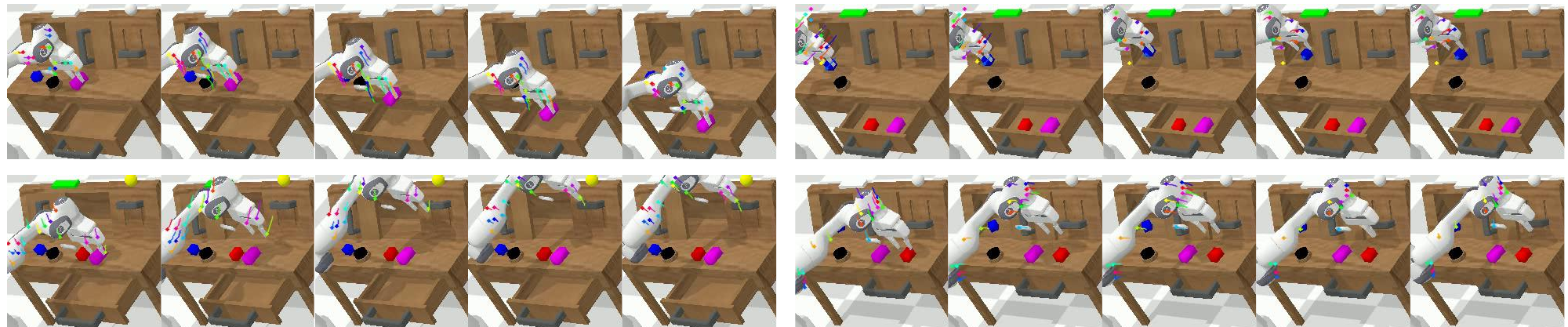}}};
    \end{tikzpicture}
  }
  % \vspace*{-0.05in}
  \caption{
    \textbf{Track annotation results on CALVIN dataset}. We select 4 sequences for visualization. 
    \textit{Best viewed when zoomed in!}
  }
 \label{fig:appTrackCalvin}
\end{figure*}

We provide qualitative demonstrations of the annotated visual tracks in Figure \ref{fig:appTrackLibero} (LIBERO) and Figure \ref{fig:appTrackCalvin} (CALVIN).

\subsection{VLA Prompt}
Given a task such as:
\begin{quote}
\textbf{[TASK]}
\textit{pick up the black bowl between the plate and the ramekin and place it on the plate}
\end{quote}
The instruction-following prompt $\task$ and track-following prompt $\taskTrack$ are:
\begin{quote}
    \textbf{[Instruction-following Prompt]}
        \textit{What 8 step actions should the robot take to [TASK] }
    \textbf{[Track-following Prompt]}
        \textit{The visual trajectory depicts a way to [TASK]. What 8 step actions should the robot take to follow it?}  
\end{quote}

\subsection{Training Configuration Details}
\label{sec:AppendixTrainDetail}

\paragraph{\method training configurations}
We show \method training configurations for all experiments in Table \ref{tab:AppendixOursTrainDetails}.
\begin{table}[h!]
    \centering
    \setlength{\tabcolsep}{3.5pt}
    \renewcommand{\arraystretch}{1.15}
    \begin{NiceTabular}{llccccccl}
        \toprule
        \multicolumn{2}{c}{\textbf{Experiment}} &
        \textbf{\makecell{Optimizer}} &
        \textbf{\makecell{Scheduler}} &
        \textbf{\makecell{LoRA\\Rank}} &
        \textbf{\makecell{Global\\Batch Size}} &
        \textbf{\makecell{GPU$^\dagger$\\Count}} &
        \textbf{\makecell{Learning\\Rate}} &
        \textbf{\makecell{Total\\Steps/Epochs}} \\
        % \cmidrule(lr){1-2}
        % \textbf{Benchmark} & \textbf{Stage} \\
        \midrule
        \Block{3-1}{LIBERO}
        & Stage 0 & Adam & Constant & 32 & 128 & 8 & 5e-4 & \makecell{60K Steps (others)\\80K Steps (Long)} \\
        \Hline
        & Stage 1 & Adam & Constant & 32 & 16 & 4 & 1e-5 & \makecell{60K Steps} \\
        \Hline
        & Stage 2 & Adam & Constant & 32 & 96 & 8 & 5e-4 & \makecell{20K Steps (Spatial, Goal)\\30K Steps (Object, Long)} \\

        \midrule
        \Block{3-1}{CALVIN}
        & Stage 0 & Adam & Constant & 32 & 128 & 8 & 5e-4 & 200K Steps \\
        \Hline
        & Stage 1 & Adam & Constant & 32 & 48 & 8 & 1e-5 & 5 Epochs \\
        \Hline
        & Stage 2 & Adam & Constant & 32 & \makecell{96 (\method)\\64 (\method-OFT)} & 8 & 5e-4 & \makecell{12 Epochs (\method)\\6 Epochs (\method-OFT)} \\
        
        \bottomrule
    \end{NiceTabular}
    \caption{\textbf{\method Training Configurations} . $^\dagger$ We use NVIDIA A100 for all of our training experiments.}
    \label{tab:AppendixOursTrainDetails}
\end{table}

\paragraph{Baseline OpenVLA/OpenVLA-OFT reproduction details}
For 8-Step OpenVLA \cite{openvla}, we use the \method Stage 1 model in all experiments since they are equivalent.
For OpenVLA-OFT \cite{openvla-oft}, which is also 8-step, we tabulate the LIBERO results directly from the original paper. For CALVIN, we train OpenVLA-OFT for 6 epochs with global batch size of 64. We observe that \textit{OpenVLA-OFT does not benefit from longer training}. When increasing the training epochs from 6 to 12, OpenVLA-OFT performance in CALVIN ABC$\rightarrow$D drops from 3.868 to 3.861.

\paragraph{Alternative training strategies details.}
Here we summarize the training details of the alternative approaches shown in Figure \ref{fig:AnalysisAblate} (B).
For DPO-then-SFT, we follow the same configuration as \method in LIBERO-Spatial.
For Co-SFT, we train the model for 100K steps, which is 40K steps more than the OpenVLA to accommodate the auxiliary task and additional training data.

\section{Additional Results}
\subsection{LIBERO Results without Receding Horizon Control}
\label{sec:AppendixLIBEROResultsFull}
\begin{table*}[h!]
    \centering
    % \small
    % \setlength\tabcolsep{2.5pt}
    \setlength\extrarowheight{1.5pt}
    \begin{NiceTabular}{lc|cccc|c}[
            % vlines = {3,7},     % vertical lines only before 2nd & 6th cell
            % hlines            % no horizontal lines
        ]
        \toprule[1.2pt]
        Methods & \makecell{Plan-Execute\\Horizon} & Spatial & Object & Goal & Long & Mean \\
        \hline
        \Block{2-1}{OpenVLA} & 
            8-8 & 73 & 84.2 & 72 & 52.8 & 70.5 \\
            & 8-4 & 79.6 & 88.6 & 82.4 & 54.4 & 76.3 \\
        
        \hline
        
        \Block{2-1}{\method} & 
        8-8
            & \makecell[t]{75.8\\ \tableGreenNote{(+2.8)}}
            & \makecell[t]{87.6\\ \tableGreenNote{(+3.4)}}
            & \makecell[t]{81.4\\ \tableGreenNote{(+9.4)}}
            & \makecell[t]{49.4\\ \tableRedNote{(-3.4)}}
            & \makecell[t]{\textbf{73.5}\\ \tableGreenNote{(+3)}} \\
        & 8-4
            & \makecell[t]{84.8\\ \tableGreenNote{(+5.2)}}
            & \makecell[t]{91.6\\ \tableGreenNote{(+3.0)}}
            & \makecell[t]{84.2\\ \tableGreenNote{(+1.8)}}
            & \makecell[t]{57.0\\ \tableGreenNote{(+2.6)}}
            & \makecell[t]{\textbf{79.4}\\ \tableGreenNote{(+3.1)}} \\
        \bottomrule[1.2pt]
    \end{NiceTabular}
    \captionsetup{width=0.7\linewidth}
    \caption{
        \textbf{LIBERO Results without Receding Horizon Control.}
        \textcolor{softgreen}{Green} (\textcolor{softred}{red}) numbers indicate performance gains (drops) relative to the OpenVLA counterpart.
    }
    \label{tab:AppendixLIBEROResultsFull}
    \vspace{-0.2in}
\end{table*}

We show the performance of OpenVLA and \method with and without the Receding Horizon Control (RHC) in Table \ref{tab:AppendixLIBEROResultsFull}.
In both settings, \method improves OpenVLA by at least $3\%$ in success rate, which futher proves the effectiveness of our approach.
In addition, removing RHC leads to a performance drop for both \method and OpenVLA in LIBERO. 
This degradation is potentially due to the declining visual conditioning when predicting later action tokens as illusrated in Figure \ref{fig:VCStudy}, which suggests that the later tokens are less reliable.
In that regime, action prediction becomes more dominated by the action distribution prior learnt from the dataset, which makes them less grounded on the world observations.
However, we do not observe performance gain by adding RHC in CALVIN. A plausible explanation is that the CALVIN training dataset is substantially larger than LIBERO, which may mitigate overfitting to action priors and support more stable long-horizon action prediction.

\subsection{Visual conditioning in other LIBERO suites}

\begin{figure*}[h!]
  % \vspace*{-0.1in}
  \centering
  \scalebox{0.85}{
    \begin{tikzpicture}
     \node[anchor=north west] at (0in,0in)
      {{\includegraphics[width=1.0\textwidth,clip=true,trim=0
      262 2 0]{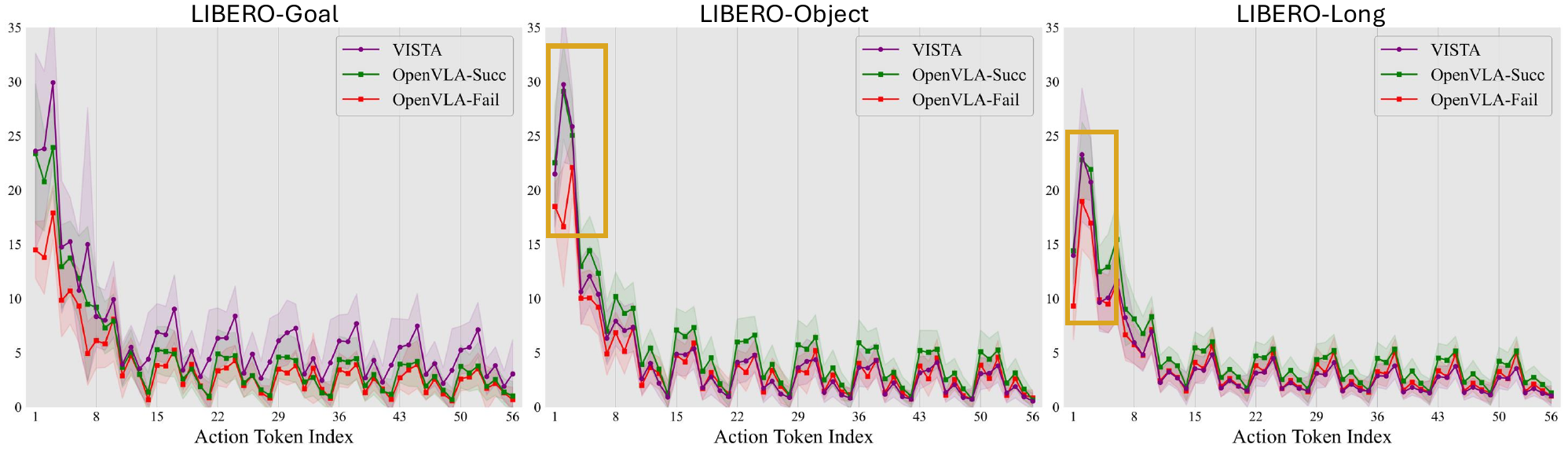}}};
    \end{tikzpicture}
  }
  % \vspace*{-0.05in}
  \caption{
    \textbf{Visual Conditioning of \method and OpenVLA in more LIBERO Suites}.
    Left: LIBERO-Goal;
    Middle: LIBERO-Object;
    Right: LIBERO-Long.
  }
 \label{fig:appVCLIBERO}
\end{figure*}

In addition to Figure \ref{fig:VCStudy}, we illustrate our quantified visual conditioning of OpenVLA and \method model in the remaining LIBERO suites in Figure \ref{fig:appVCLIBERO} following the protocol in Sec. \ref{sec:VCDefine}.
We make two observations from the plots.
First, the successful trials of OpenVLA constantly feature stronger visual grounding compared to failure ones across all suites, which further support our claim on the importance of visual clue utilization to the performance of VLA.
Second, \method improves the visual conditioning of all action tokens in LIBERO-Goal as in LIBERO-Spatial.
As for LIBERO-Object and LIBERO-Long, although the visual conditioning is not improved in the prediction of later tokens, it is enhanced to the level of the average successful OpenVLA rollouts at the early stage of the prediction as highlighed by the goal bounding box. 
Coupled with Receding Horizon Control (RHC), the performance is improved as presented in Table \ref{tab:ResultsLIBERO}.
However, the inability to raise visual conditioning at the later stage might explain the degraded performance of \method without RHC in Table \ref{tab:AppendixLIBEROResultsFull}, which is left for future work.

\subsection{Ablation on the Distillation Loss Weight}
We ablate on the distillation loss weight $\distillWeight$ on LIBERO-Spatial. 
We experiment on $\distillWeight \in [0.01, 0.1, 1]$ in \method Stage 3 training, where all training configurations are the same as the ones tabulated in Table \ref{tab:AppendixOursTrainDetails}, except that we reduce the global batch size from 96 to 48 to reduce the required GPU from 8 to 4 to save compute.
We also evaluate on 200 trails instead of 500 as in the main paper. 
As shown in Table \ref{tab:AppendixDistillWeightAblation}, althought \method with all $\distillWeight$ outperforms OpenVLA (presented in Table \ref{tab:AppendixLIBEROResultsFull}), $\distillWeight=0.1$ achieves the best result. Hence we fix $\distillWeight=0.1$ in all main experiments.

\begin{table*}[h!]
    \centering
    % \small
    \setlength\tabcolsep{10pt}
    \setlength\extrarowheight{1pt}
    \begin{NiceTabular}{cccc}[
            % vlines = {3,7},     % vertical lines only before 2nd & 6th cell
            % hlines            % no horizontal lines
        ]
        \toprule[1.2pt]
        \Block{2-1}{\makecell{Plan-Execute\\Horizon}} & \Block{1-3}{$\distillWeight$} \\
        \Hline
        & 0.01 & 0.1 & 1 \\
        \midrule
        8-8 & 72 & \textbf{76} & 73 \\
        8-4 & 81.5 & \textbf{86} & 85 \\
        \bottomrule[1.2pt]
    \end{NiceTabular}
    \captionsetup{width=0.5\linewidth}
    \caption{
        \textbf{LIBERO-Spatial Success Rate (\%) under Different Distillation Loss Weight $\distillWeight$.}
    }
    \label{tab:AppendixDistillWeightAblation}
    % \vspace{-0.2in}
\end{table*}

% \input{icml2026/sec_math}

%%%%%%%%%%%%%%%%%%%%%%%%%%%%%%%%%%%%%%%%%%%%%%%%%%%%%%%%%%%%%%%%%%%%%%%%%%%%%%%
%%%%%%%%%%%%%%%%%%%%%%%%%%%%%%%%%%%%%%%%%%%%%%%%%%%%%%%%%%%%%%%%%%%%%%%%%%%%%%%

\end{document}